\documentclass[10pt]{article}

\usepackage[margin=1.3in]{geometry}  
\usepackage[parfill]{parskip}  
\usepackage{graphicx}
\usepackage{amsmath,amssymb,amsthm}
\usepackage[algoruled]{algorithm2e}
\usepackage{epstopdf}
\usepackage{xifthen}
\usepackage{enumitem}
\usepackage[colorlinks]{hyperref}
\usepackage{todonotes}
\usepackage{mathtools}
\usepackage[displaymath, mathlines]{lineno}
\usepackage{bm}

\usepackage{fancyhdr}
\usepackage{sectsty}
\sectionfont{\fontsize{10}{15}\selectfont}
\subsectionfont{\fontsize{10}{15}\selectfont}
\subsubsectionfont{\fontsize{10}{15}\selectfont}

\def\vx{{\bm{x}}}

\def\mC{{\bm{C}}}

\DeclareMathAlphabet{\mathsfit}{\encodingdefault}{\sfdefault}{m}{sl}
\SetMathAlphabet{\mathsfit}{bold}{\encodingdefault}{\sfdefault}{bx}{n}

\def\gH{{\mathcal{H}}}
\def\gI{{\mathcal{I}}}

\def\gL{{\mathcal{L}}}
\def\gM{{\mathcal{M}}}
\def\gN{{\mathcal{N}}}

\def\gS{{\mathcal{S}}}

\def\sN{{\mathbb{N}}}

\def\sR{{\mathbb{R}}}

\usepackage{amsthm}
\usepackage{amsfonts}
\usepackage{amsmath,bm}

\newtheorem{mydef}{Definition}

\newtheorem{myrem}{Remark}
\newtheorem{myex}{Example}

\newcommand{\norm}[1]{\left\lVert#1\right\rVert}

\usepackage{xargs}                      % Use more than one optional parameter in a new commands
\usepackage{mathrsfs}
\newcommand{\St}{ \mathscr{S}}

\newcommand{\Sti}{ \stackrel{\circ}{\mathbf{\St}}}

\newcommand{\Stc}{ \partial\mathscr{S}}

\newcommand{\Ca}{\bm{C}}

\newcommand{\lc}{\left\{}
\newcommand{\rc}{\right\}}

\hypersetup{
    colorlinks=true,
    linkcolor=blue,
    filecolor=magenta,      
    urlcolor=cyan,
    pdftitle={Overleaf Example},
    pdfpagemode=FullScreen,
    citecolor = blue,
    }

\usepackage{subfigure}

\usepackage[style=alphabetic,backend=bibtex,maxbibnames=99]{biblatex}
\usepackage{caption}

\title{Towards Modeling and Resolving Singular Parameter Spaces using Stratifolds\footnote{A preliminary version of this work was presented at NeurIPS 2021 as a Spotlight in the \href{https://opt-ml.org/index.html}{13th Annual Workshop on Optimization for Machine Learning (OPT2021)}}}

\author{Pascal Mattia Esser\footnote{Technical University of Munich, Germany},
\qquad
Frank Nielsen\footnote{Sony Computer Science Laboratories Inc., Japan}}

\begin{document}

\maketitle

\begin{abstract}%
When analyzing parametric statistical models, a useful approach consists in modeling geometrically the parameter space. However, even for very simple and commonly used hierarchical models like statistical mixtures or stochastic deep neural networks, the smoothness assumption of manifolds is violated at singular points which exhibit non-smooth neighborhoods in the parameter space. These singular models have been analyzed in the context of learning dynamics, where singularities can act as attractors on the learning trajectory and, therefore, negatively influence the convergence speed of models. We propose a general approach to circumvent the problem arising from singularities by using stratifolds, a concept from algebraic topology, to formally model singular parameter spaces. We use the property that specific stratifolds are equipped with a resolution method to construct a smooth manifold approximation of the singular space. We empirically show that using (natural) gradient descent on the smooth manifold approximation instead of the singular space allows us to avoid the attractor behavior and therefore improve the convergence speed in learning.
\end{abstract}

\section{Introduction}
\label{sec: Introduction}

A \emph{parameter space} is a subspace of the Euclidean space equipped with the metric topology induced by the $L_2$-norm that includes all permissible parameters of a statistical model. This geometric viewpoint of the set of parameters allows us to analyze how specific topological properties given by the parameter space influence the learning dynamics. Specifically, regular parameter spaces are modeled  as Fisher-Rao manifolds \cite{Hotelling-1930} to allow   continuous and smooth update.  However, even commonly used simple models like neural networks \cite{sun2021lightlike} violate the manifold assumption. We will refer to such points in the parameter space that are \emph{continuous but not smooth} as \emph{singular}.
This setting results into  two main problems:
\begin{enumerate}
    \item  The tangent space on the parameter space is not well defined  \emph{at the singularity} due to the lack of smoothness. Therefore, gradient-based learning algorithms in a gradient flow setting fail here.
\item  Several studies \cite{SingularDynamics,SingularDynamics2,SingularDynamica3} have shown that \emph{near the singularity} an attractor behavior \cite{attractor} can be observed such that update steps gets smaller close to the singularity.
\end{enumerate}
\clearpage
 The latter one is due to the fact that the Hessian of the loss function becomes singular (also referred to as degenerate or higher-order saddles) when the parameters space is singular, leading to slower convergence when training the model \cite{highOrderSaddle, watanabe_2009}.

\emph{To overcome those problems we propose to model the parameter space as a specific topological space, namely Stratifolds \cite{Toshiki2016category,StratifoldsBook,ResolutionOS} and then use a resolution of the space to obtain a smooth manifold approximation around the singularities.}

\begin{figure}[t!]
% \floatconts
%   {fig: cusp}% label for whole figure
    \subfigure{%
      \label{fig: cusp model}% label for this sub-figure
      \includegraphics[width=0.32\textwidth]{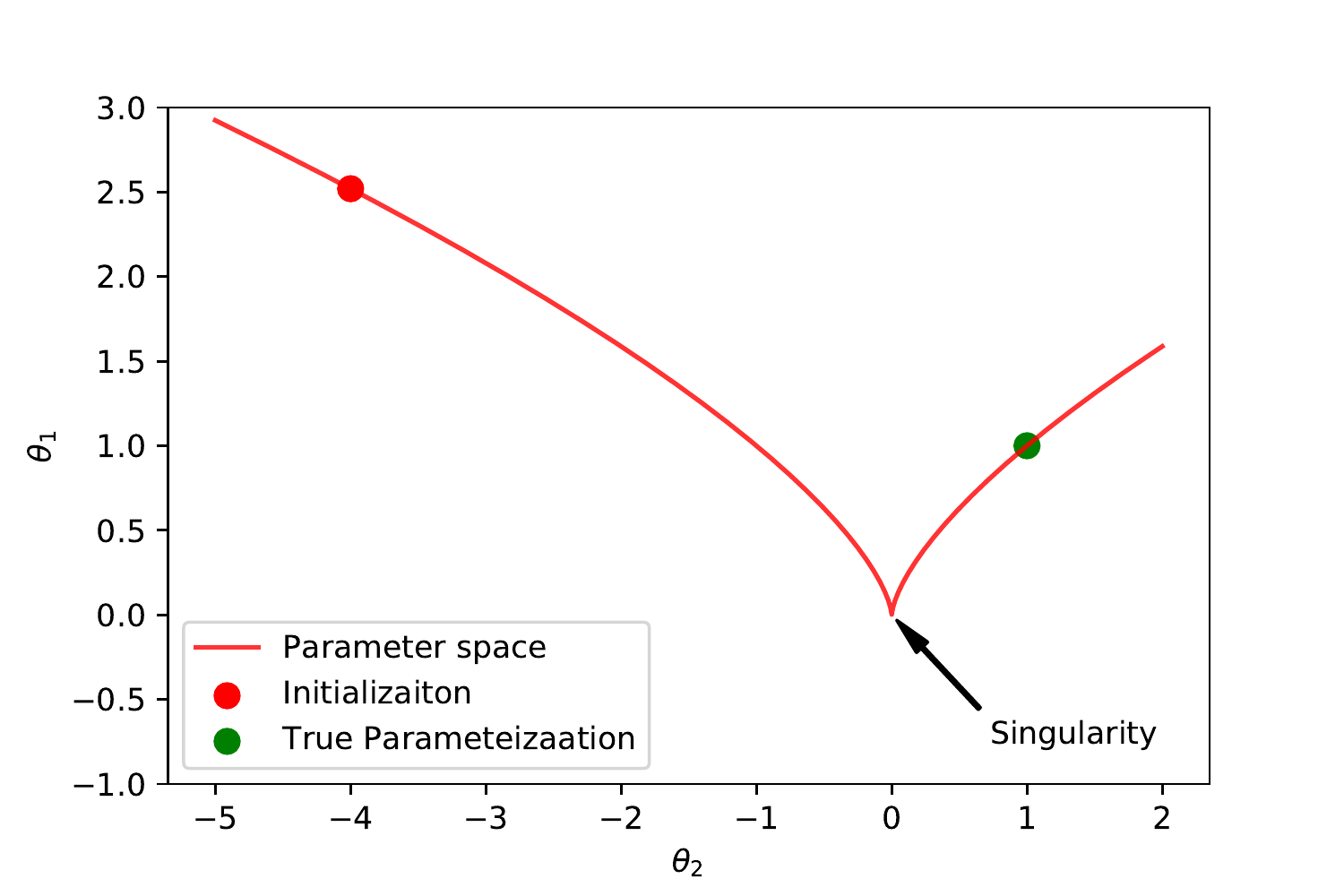}
    } % space out the images a bit
    \subfigure{%
      \label{fig: cusp approx}% label for this sub-figure
      \includegraphics[width=0.32\textwidth]{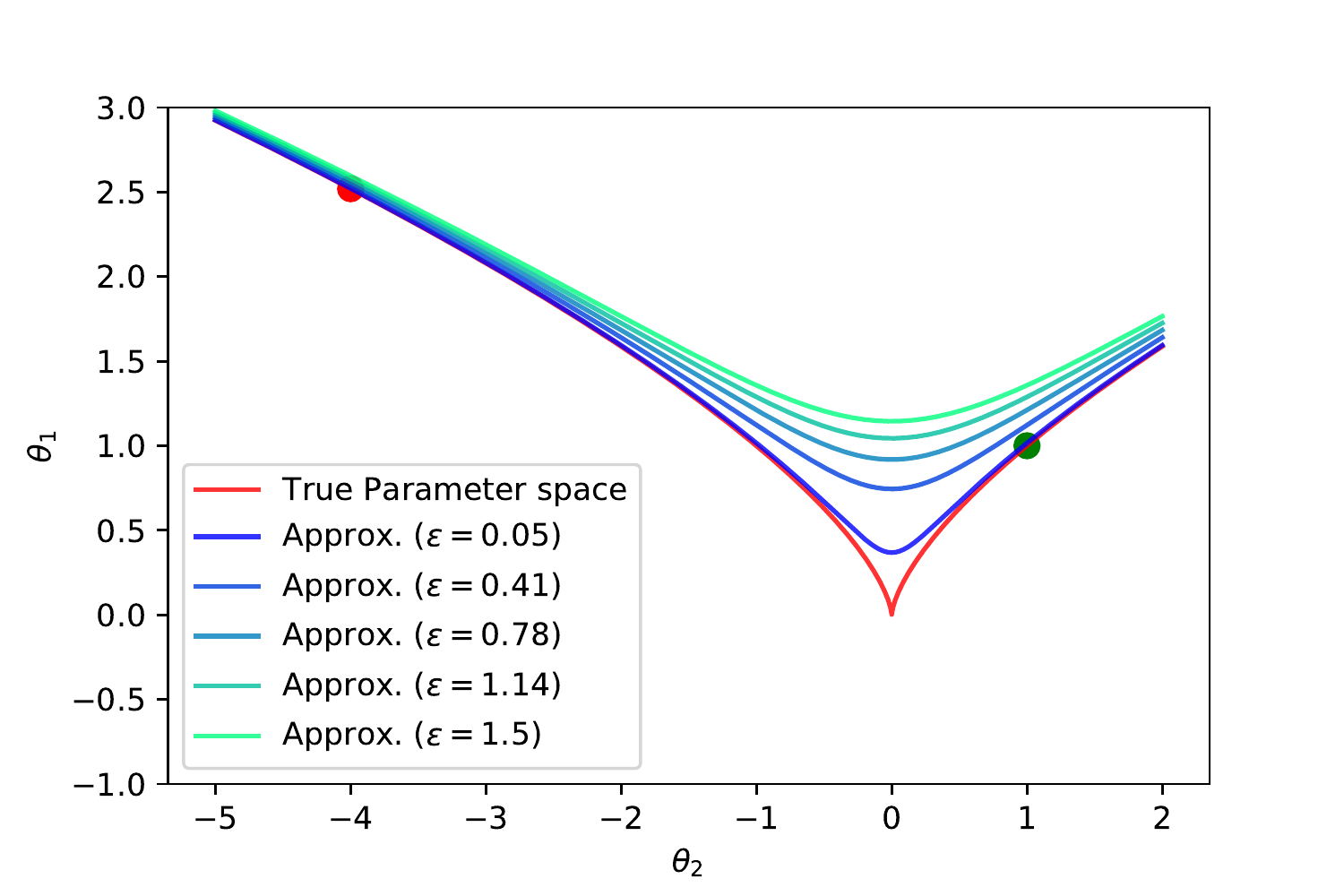}
    }
     \subfigure{%
      \label{fig: cusp gradient}% label for this sub-figure
      \includegraphics[width=0.32\textwidth]{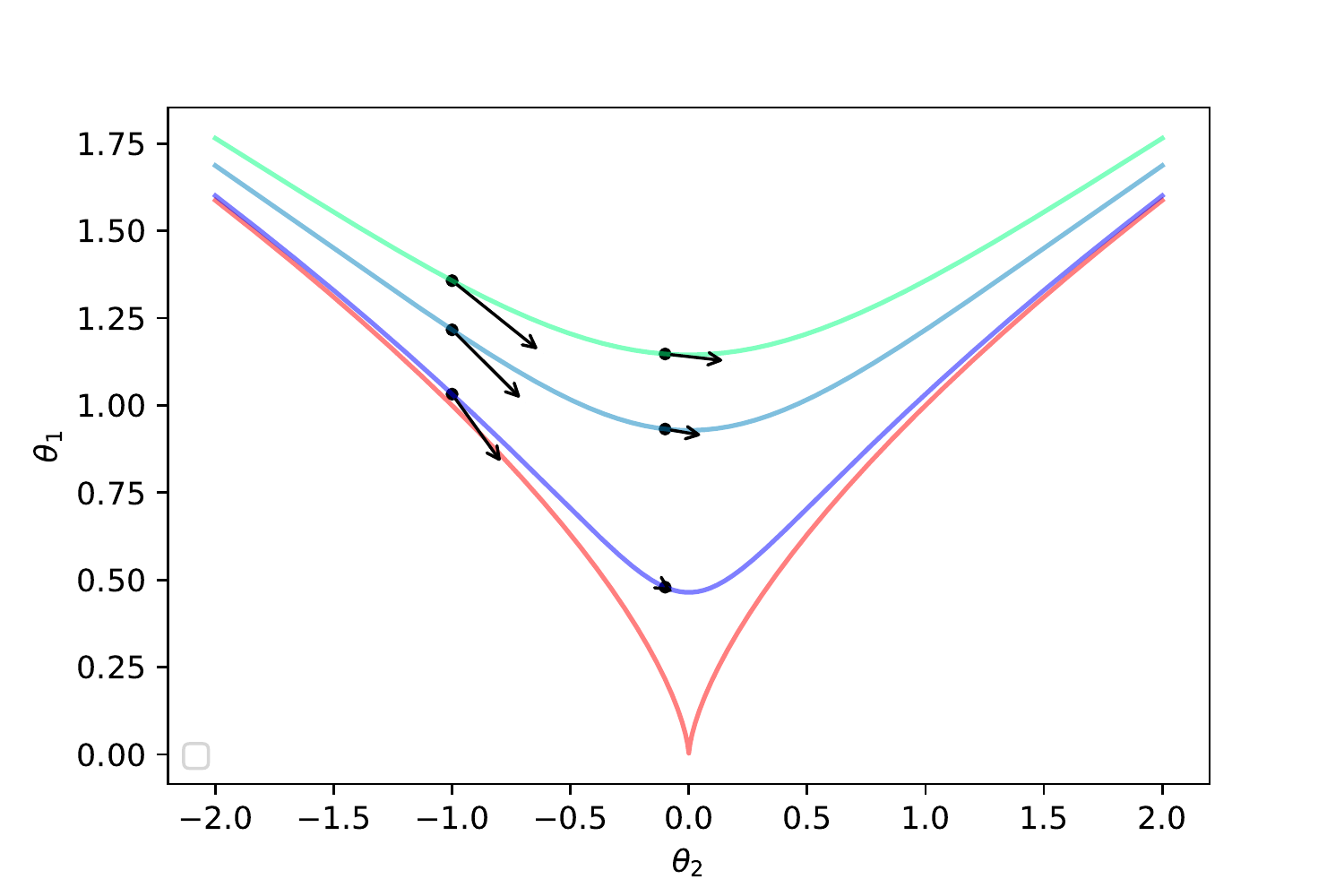}
    }
    \caption{
  (a) Singular parameter space in red. Initialization at the red point. True parameterization at the green point. The update has to go through the singularity.
  (b) Proposed smooth approximation.
  (c) Gradients at different points on the singular parameter space and the smooth approximations.
  }
% \subfigure
 \label{fig: cusp}
\end{figure}

We illustrate the main idea on a specific example shown in Figure~\ref{fig: cusp}. Consider a function $f_{\theta_1,\theta_2}(x)$ such that the parameters lie on a topological space restricted to $\theta_1^2+\theta_2^3=0$ as displayed  in Figure~\ref{fig: cusp} (a). A standard learning problem is now to start at some initialization (shown by the red dot) and try to reach a true parameterization (shown by the green dot) using a gradient decent (GD) approach. We note that at the singularity at $(\theta_1,\theta_2)=(0,0)$ the space has a singularity, characterized by the lack of a unique tangent space.
We can formally model such a space as a Stratifold by decomposing it into a set of smooth manifolds.
To overcome the problem arising from the lack of smoothness we propose to instead consider a manifold approximation of the singular topological space as illustrated in Figure~\ref{fig: cusp} (b), which can be formalized as  $f_{\theta_1,\theta_2,\epsilon}(x)\text{ s.t. }\theta_1^2+\theta_2^3=\epsilon $. In the main part of the paper we will formalize this idea. Depending on how close the approximation is, we observe the following tradeoff: for a close approximation (in this case small $\epsilon$) even points close to the singularity are well approximated however, as mentioned above, singularities admit an attractor behavior, which is stronger the closer the approximation is to the original model. We can see this illustrated in Figure~\ref{fig: cusp} (c). Away from the singularity differences in  gradients are very similar while close to the singularity the gradients on the approximation are large, allowing for the effective application of GD, while they degenerate near the singularity.

\begin{figure}[t]
    \centering
    \includegraphics[width = 0.8\linewidth]{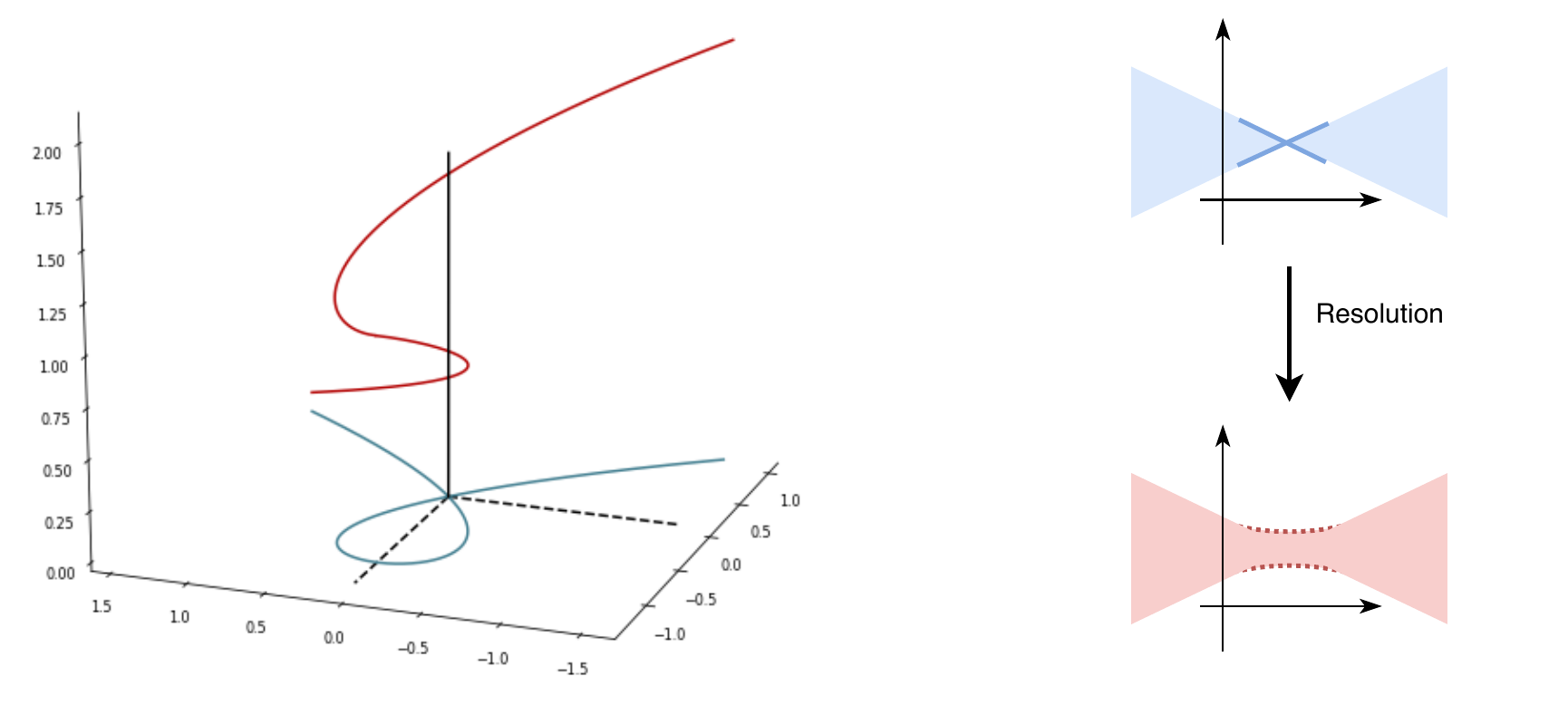}
    \caption{Illustration of the resolution of singular points.
    \emph{(left)} Blowup following \cite{watanabe_2009}. This illustrates a functional approach where the function is blown up into a higher dimension.
    \emph{(right)} Our approach. Preserving the dimension, we propose to construct a smooth manifold approximation of the singular topological parameter space.
    }
    \label{fig:blowup}
\end{figure}

While there are a few works that consider Stratifolds or similar constructions in the context of learning theory \cite{sun2021lightlike,boissonnat2020} and focus on local manifold structures \cite{Lin2021ICML}, to the best of our knowledge this is the first approach to consider Stratifolds and their resolution in the context of learning dynamics.
Furthermore there are several approaches on the analysis of specific singular models where the most general one, based on a topological viewpoint, is by \cite{watanabe_2009,Watanabe2010,Watanabe2013}. 
While  \cite{watanabe_2009} relies in principle on the same resolution theorems for algebraic varieties \cite{ResolutionAlgebraicVarietyI,ResolutionAlgebraicVarietyII} as we do, \cite{watanabe_2009} focuses on information criteria,  as standard approaches  \cite{BIC,AIC} do not hold in the singular regime. The technique of~\cite{watanabe_2009} uses a \emph{blowup} of the function space into higher dimensions and therefore follows a \emph{functional perspective} where in contrast we consider a resolution that preserves the dimension of the manifold, motivated by a \emph{topological viewpoint}. We illustrate this idea in Figure~\ref{fig:blowup}

The remainder of the paper is structured as follows. We start with Section~\ref{sec: stratifold construction} and Section~\ref{sec resolution of singularity} to introduce the construction and resolution of Stratifolds which are then applied to a simple toy model in Section~\ref{sec: cone model}. From there we illustrate that the above presented idea is relevant to the machine learning community by showing how using the resolution avoids the attractor behavior and improves the learning convergence rate.

\section{Introduction to Stratifolds}\label{sec: Introduction to stratifolds}

Equipped with the above-given intuition, we can now define the central concept for this paper: \emph{Stratifolds}. To do so, we follow \cite{StratifoldsBook, ResolutionOS, Toshiki2016category} but restrict ourselves to outlining the main ideas. We provide further definitions and details in Appendix~\ref{app: sec: additional definitions 1} and \ref{app: sec: additional definitions 2}.

Let $\mC \subset C^0(\St)$ be a locally detectable subalgebra then we can describe the decomposition for a differential space $(\St, \Ca)$ over the subspace $\St^i:=\{x\in\St|\ \dim (T_x\St) = i\}$ and the disjoint union  $\St = \bigsqcup_i \St^i$.
Furthermore we introduce the following terminology:
Let $\St^i$ be the $i$-stratum of $\St$ and let $\bigcup_{i\leq r}\St^i = : \Sigma^r$ be the $r$-skeleton of $\St$.
Now a $n$-dimensional Stratifold $\St$ is a topological space $\St$ together with a class of distinguished continuous (smooth) functions $\St\rightarrow\sR$. This generalizes smooth manifolds $\mathcal{M}$ where the class of considered functions are in $\Ca^\infty$. An $n$-dimensional Stratifold is a smooth manifold iff $\St^i = \emptyset,~\forall\ i<n$. For the Stratifold, this class of smooth functions offers the decomposition.
A natural example of spaces with singularities occur in algebraic geometry as algebraic varieties, i.e., zero sets of a family of polynomials.

	\begin{figure}[t!]
	\centering
	\includegraphics[width=0.9\linewidth]{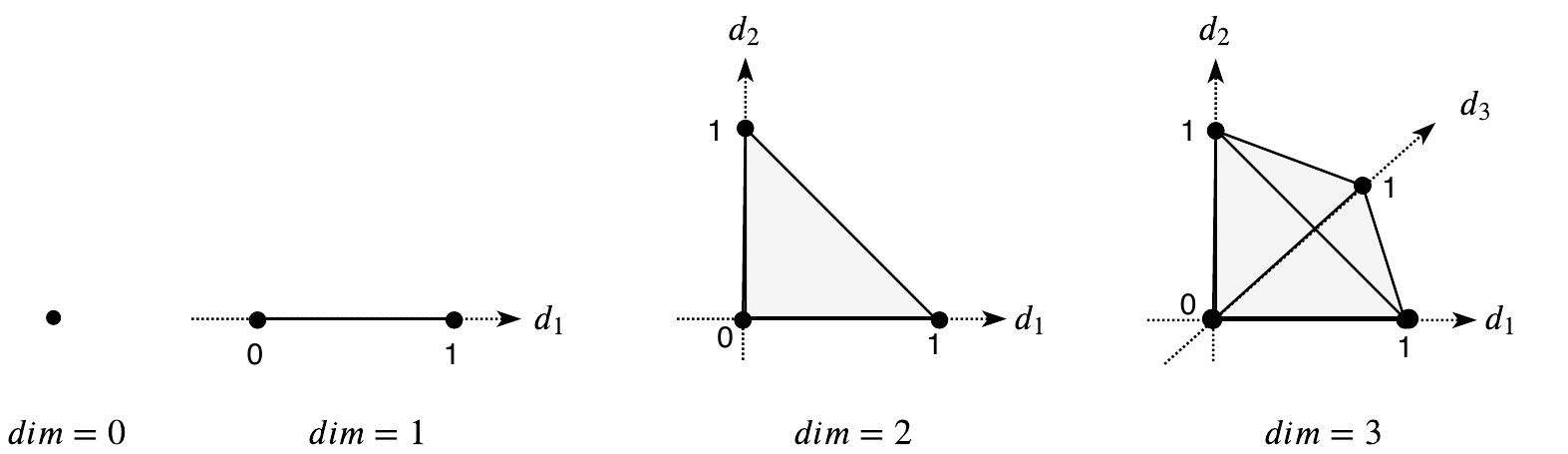}
	\caption[Illustration of a Probability Simplex]{Probability simplex in a $\{d_1, d_2, d_3\}$ coordinate system: 0-simplex (point), 1-simplex (line segment), 2-simplex (triangle), 3-simplex (tetrahedron).}
	\label{fig: simplex}
\end{figure}%

\subsection{Inductive Construction of a Stratifold}\label{sec: stratifold construction}

Given the concept of Stratifolds, an immediate question that comes up is \emph{given a parameter space, how can we construct a Stratifold that describes it?}

We consider an inductive construction, as defined in \cite{StratifoldsBook}, where we start with the lowest dimensional stratum and then by induction iteratively glue higher dimensional strata on it until reaching the top stratum. 
Start by considering a $n$
 dimensional Stratifold $(\St,\Ca)$ and a smooth manifold $\mathcal{W}$ together with a boundary with a collar $\bm c:\partial\mathcal{W}\times [0,\epsilon)\rightarrow\mathcal{W}$ and furthermore assume $k>n$. In addition we define the morphism $f:\partial\mathcal{W}\rightarrow\St$ to be the \textit{attaching map}. 
 Now we define $\St'$ by gluing $\mathcal{W}$ to $\St$ with $f$ by:
$
     \St':=\mathcal{W}\cup_f\St
$.
 On this space we consider the algebra $\Ca'$ consisting of functions $g : \St^{\prime} \rightarrow \mathbb{R}$ whose restriction to $\St$ is in $\Ca$ whose restriction to the interior of $\mathcal W, ~ \stackrel{\circ}{\mathcal W} :=W-\partial\mathcal W$  is smooth and such that for some $\delta<\epsilon$ we have $g\, \mathbf{c}(x, t)=g\, f(x)$ for all $x \in \partial\mathcal W \text { and } t<\delta$.

To illustrate this idea, we can now apply the construction as described above on the example of a probability simplex as shown in Figure~\ref{fig: simplex}.
Consider the specific case of a simplex where the simplex vertices $u_{0}, \dots, u_{k} \in \mathbb{R}^{k}$ are $k+1$ standard unit vectors forming the \emph{Standard $n$-simplex}, denoted by $\Delta^{n}$.
An example of use of the probability simplex in practice is the parameter space of  \emph{multinomial distributions}: for $n$ independent trials and $k$ possible mutually exclusive outcomes, with corresponding probabilities $p_{1}, \ldots, p_{k}  \in \mathbb{R}^{k}\ |\ \sum_{i=0}^{k} p_{i}=1 ,p_{i} \geq 0\ \forall i\in[k]$, such that $p_i$ lies on the simplex.
We can now use the previously define inductive construction to obtain a Stratifold formulation.
 We have trivially for $\Delta^0$ a $0$-dimensional manifold and therefore also a $0$-dimensional Stratifold: $\St^0 =\Delta^{0}=\{\left(t_{0}\right) \in \mathbb{R}^{1}\ |\ \sum_{i=0}^{0} t_{i}=1 \}$ which we  use as a starting point and from there glue higher dimensional simplices on it. 
 In addition the algebra $\Ca$ is all constant functions on $\St^0\in \mathbb{R}^{1}$ .
 ~
 Therefore in the spirit of induction, we can consider the $\Delta^0$ and the gluing of $\Delta^1$ on it as a kind of base case.
 ~
 Now consider $\Stc = \St - \Sti$, where we define $\Stc = \Delta^{0}$ and
 $\St = \Delta^{1}=\{\left(t_{0}, t_{1}\right) \in \mathbb{R}^{2}\ |\ \sum_{i=0}^{1} t_{i}=1\}$ and 
 $\Sti = \Delta^{1}-\Delta^0=\{\left(t_{0}, t_{1}\right) \in \mathbb{R}^{2} | \sum_{i=0}^{1} t_{i}<1\}$
 where $ t_{i} \geq 0\ \forall i$.
 Then the algebra $\Ca$ is all functions that are smooth on $\Sti$ and constant on $\Stc$. We can define the space by starting with initial Stratifold   $(\St, \Ca) := \Delta^0$, considering a $k>n$ dimensional manifolds with boundary, in this case $\mathcal W:=\Delta^1$. Finally the attaching map $f : \partial \mathcal W \rightarrow \St$ is an identity map as $\Delta^0 =  \partial \mathcal W$ (the lower dimensional simplex lies on the boundary of the higher dimensional one).
From there, the inductive step would be to consider the newly constructed $\Delta^1$ as a starting point and attach $\Delta^2$ onto it.

\subsection{Resolution of Stratifolds}\label{sec resolution of singularity}
% For this section, we again mainly follow \cite{ResolutionOS}.
While the previous section provided us with a framework for modeling parameter spaces that are not smooth manifolds, the descriptive element does not improve the model performance. Therefore we are interested in a \emph{resolution} of the Stratifold that provides a smooth manifold approximation leaving the dimension of the top stratum untouched.
In this paper, we consider the setting of isolated singularities. The work of~\cite{ResolutionOS} extends this concept but the details become quite technical and we therefore refer to this analysis as future work.
To do so we start with defining the setup more formally \cite{ResolutionOS}.

Let $\St$ be an $m$-dimensional Stratifold. A resolution of $\St$ is a map $p:\hat{\St}\rightarrow\St$ such that
\begin{enumerate}
    \item  $\hat{\St}$ is a smooth manifold,
    \item $p$ is a proper morphism,
    \item the restriction of $p$ to $p^{-1}(\St^m)$ is a diffeomorphism onto $\St^m$,
    \item $p^{-1}(\St^m)$ is dense in $\hat{\St}$.
\end{enumerate}
If $\St$ is an algebraic variety, \cite{ResolutionAlgebraicVarietyI,ResolutionAlgebraicVarietyII} shows that there is a resolution of the singularity using algebraic geometry. This link is especially interesting considering that the method of~\cite{watanabe_2009} is built on the same theorems but in the context of functional blowup, and not in the context of topological resolutions. In the following we will illustrate the construction and resolution on a specific example.

\begin{figure}[t]
	\centering

    \subfigure{%
      \label{fig: cone model}% label for this sub-figure
      \includegraphics[width=0.35\textwidth]{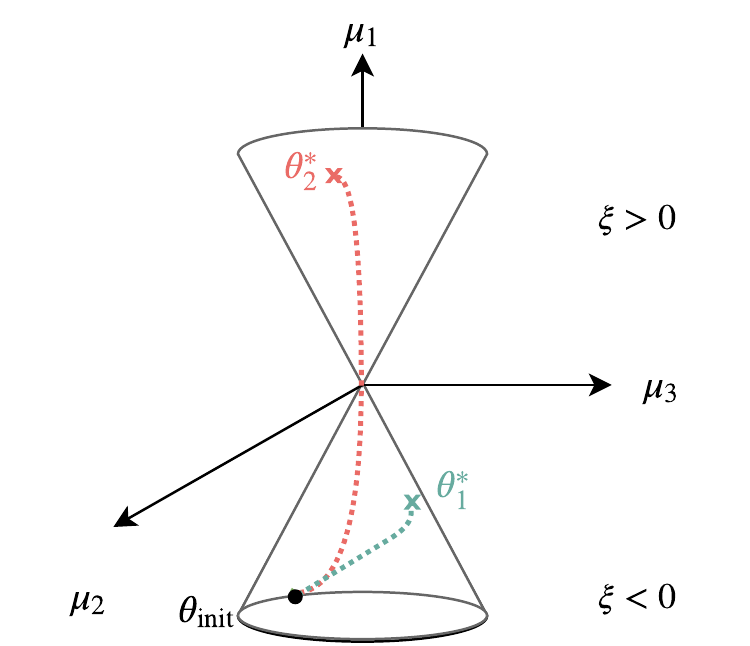}
    }\qquad % space out the images a bit
    \subfigure{%
      \label{fig: cone strat}% label for this sub-figure
      \includegraphics[width=0.55\textwidth]{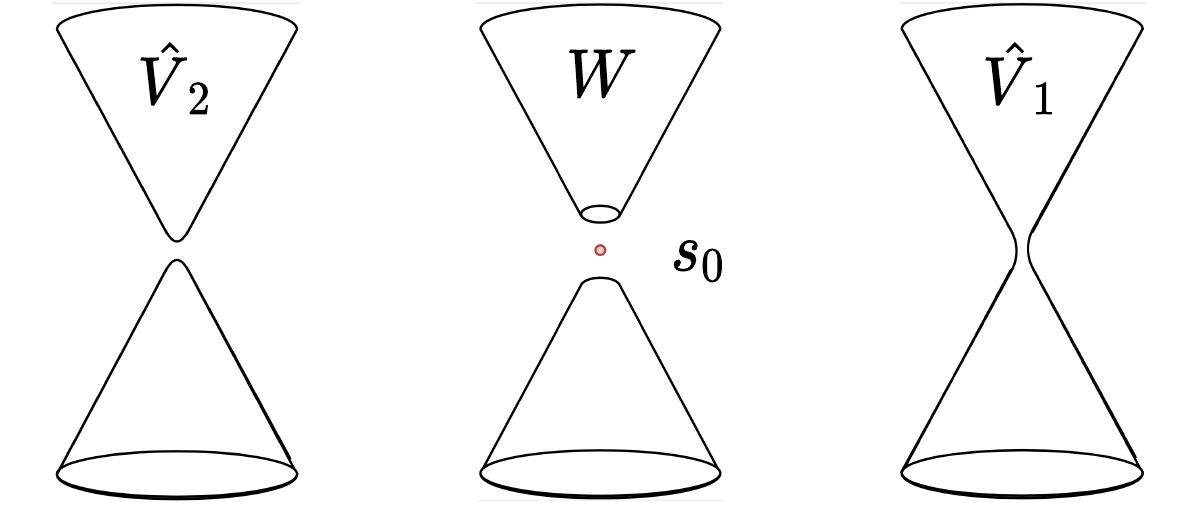}
    }
    \caption{
  (a) Parameter space of the double cone model, together with examples of learning dynamics. Let $\bm\theta_{\mathrm{init}}$ be the parameter initialization, then $\bm\theta_1^*$ is an example for a true parameterization that lies on the same cone, where as $\bm\theta_2^*$ is an example for a true parameterization that lies on the second cone.
  ~
  (b) \emph{(left)} two sheet hyperboloid. \emph{(middle)} Stratification. \emph{(right)} one sheet hyperboloid.
  }
  \label{fig: cone}
% \subfigure

\end{figure}

 \subsection{The Cone Model as a Stratifold: Inductive Construction and Resolution }\label{sec: cone model}
 
 For our analysis,  we  look at a simple model that is used to show the influence of singularities on learning trajectories and convergence speed \cite{SingularDynamics,SingularDynamics2,InitalConePaper,SingularDynamica3}.
 Consider a random variable $\bm x \in \sR^{3}$, subject to a Gaussian distribution with mean $\bm \mu$ and covariance matrix, the identity matrix. The hypothesis class over all such function is given by
$\textstyle
\gH_{\bm\theta}:=\big\{h_{\bm\theta}(\vx) =
\frac{1}{\sqrt{2\pi}^{3}}\exp{\big(-\frac{1}{2} \norm{\bm x - \bm \mu}^2 \big)}\  
\big|\ \bm\mu\in\sR^3,~ \mu_1^2 + \mu_2^2 - \mu_3^2 = 0
\big\}
$ where $\bm \mu$ is restricted to be on the surface of a double cone.

As the cone lies in $\sR^3$, we can reparameterize the cone as:
    $  
    \mu_1 = \xi, \ 
    \mu_2  = \xi\cos{\theta}, \ 
    \mu_3 = \xi\sin{\theta}
    $.
%  such that $\bm \Theta:=\{\xi,\theta\}$ parameterizes the cone model.
 This results in two cones, one for $\xi\geq0$ and one for $\xi\leq0$, connected at the apex $\xi = 0$, which we  refer to as the singularity of the model. This means that the parameter surface is given by
$ \St = \lc (\mu_1, \mu_2, \mu_3)\in \mathbb{R}^3  \ |\  \mu_1^2+\mu_2^2=\mu_3^2 \rc
$.
We can see this model illustrated in Figure~\ref{fig: cone} (a).

We are now interested in the applying the steps as described in Section~\ref{sec resolution of singularity} to define the model as a Stratifold and then also derive a resolution of it.

 Following directly from the definition, it is easy to see that we can not define a unique tangent plane on the singularity at $\bm\mu = \{0,0,0\}$. Furthermore we also see that the point is closed. We define the zero set as
$
\Sigma := \lc s_i \rc_{i=0} 
=s_0
= (0,0,0) \quad \Rightarrow \quad \text{dim}(\Sigma) = 0.
$
This gives us an algebraic variety $V\subset \mathbb{R}^3$ with one isolated singularity and in line with the earlier definitions the distance function becomes  $\rho_i(x): = ||x-s_0||^2$, which  gives a distance to the origin of the coordinate system.
The $\varepsilon$-enclosing ball is  around the origin and we can see $V_{\varepsilon_i}(s_i):=V\cap D_{\varepsilon_i}(s_i)$ as the part of the double cone that is within the $\varepsilon$-ball. We can see this property illustrated in Figure~\ref{fig: cone} (b) (middle).
Seeing ${D}_{\varepsilon_{i}}$ to be the enclosing ball, we can see $\mathring{D}_{\varepsilon_{i}}$ as the $\varepsilon$-ball without the border.
As we only have one singularity, we can define $W$ now as:
$ W :=V-  \mathring{D}_{\varepsilon_{i}}\left(s_{0}\right) \cup_{\mathrm{id}} \partial V_{\varepsilon_{i}}\left(s_{i}\right) \times\left[0, \varepsilon_{i}\right]
$. Here $W$ builds now the double cone structure away from the singularity where the enclosing ball builds the corners close to the singularity.

This gives a final Stratifold as one $0$-dimensional manifold of the singularity, $\St^0$, and the manifold of the cone, with the corner build by the enclosing ball around the singularity $\St^0$ as:
$
 \St^0 = s_0, ~ \St^2 =  W
$.
 And the complete Stratifold as
$      \St = \bigsqcup_i \St^i = \St^0 \sqcup \St^2 = s_0\sqcup   W.
$ %
For the cone model, define $h(\mu_1 , \mu_2, \mu_3) = \mu_1^2 + \mu_2^2 - \mu_3^2 = 0$ and following Section~\ref{sec resolution of singularity} we have
$\mathscr{V} = h^{-1}(0)$
with a not optimal resolution of
$\widehat {\mathscr{V}}_2 = h^{-1}(\varepsilon)$
which gives us a hyperboloid of two sheets (Figure~\ref{fig: cone} (b), left) and an optimal resolution of
$\widehat {\mathscr{V}}_1 = h^{-1}(-\varepsilon)$
which gives us a hyperboloid of one sheet as illustrated in Figure~\ref{fig: cone} (b) (right).

\section{Experiments}\label{sec: experiments}

\begin{figure}[t!]
	\centering
    \subfigure{%
      \label{fig:pic1}% label for this sub-figure
      \includegraphics[width=0.6\linewidth]{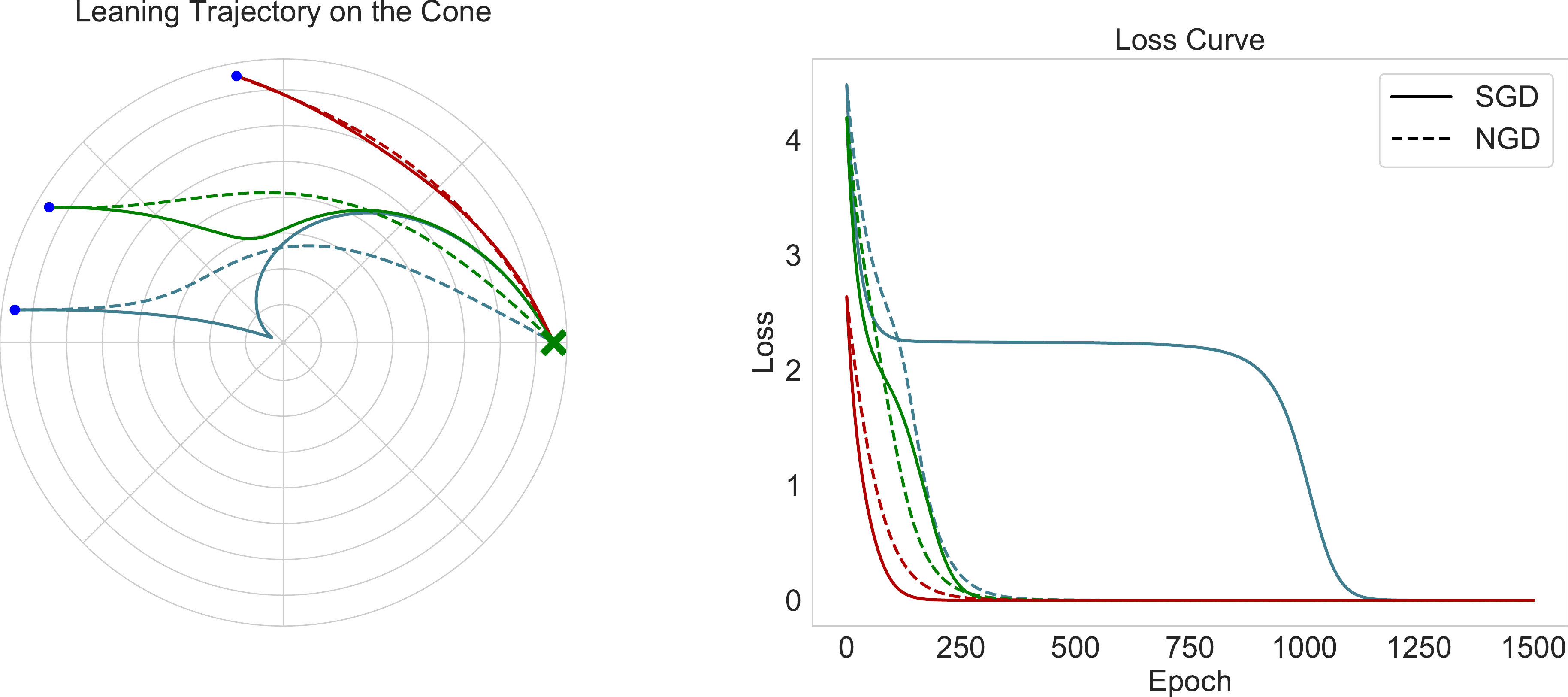}
    }\qquad % space out the images a bit
    \subfigure{%
      \label{fig:pic2}% label for this sub-figure
      	\includegraphics[width=0.6\linewidth]{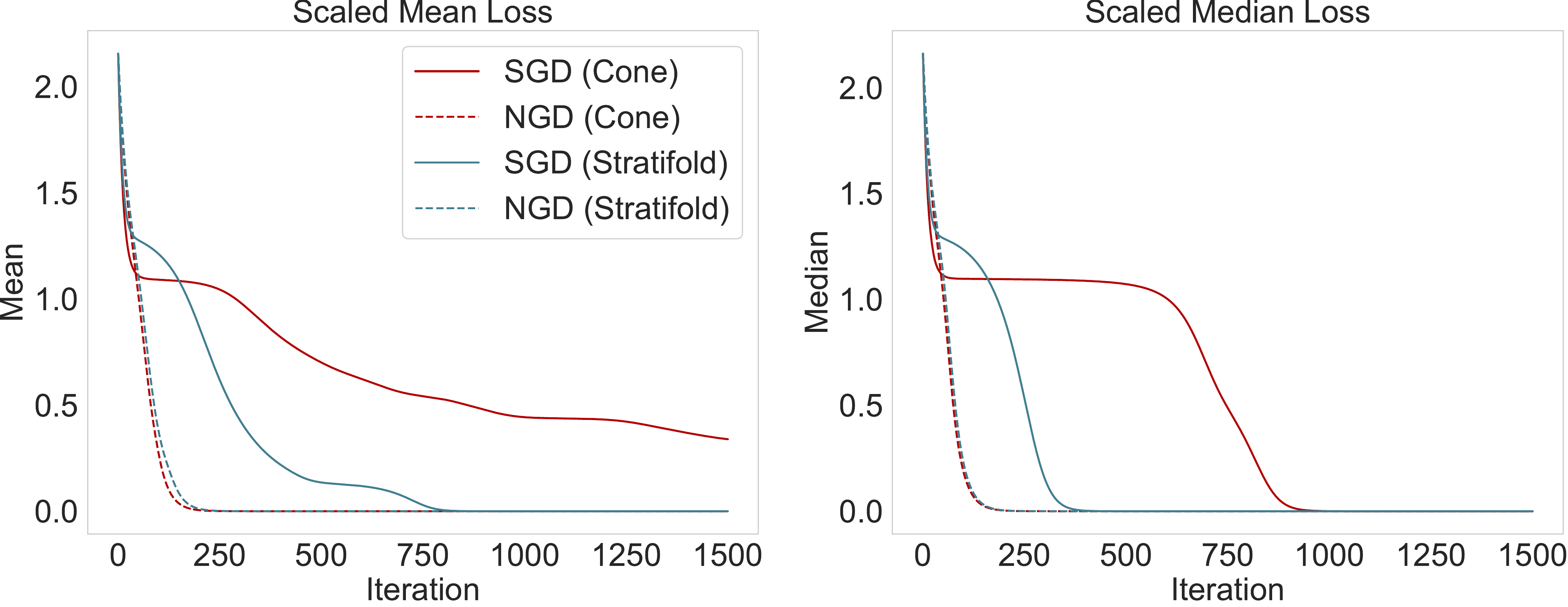}
    }
    \caption{
  (a) \emph{(left)} top view onto the cone model. Displayed are the learning trajectories for three different initializations (blue dots) for the same target value (green cross). Solid lines: learning trajectories for the GD on the original cone model. Dotted lines: learning trajectories GD on the Stratifold resolution.
	 \emph{(right)} Loss functions corresponding to the plotted learning trajectories.
  ~
  (b) Loss for NGD and SGD on the Cone Model. 
		 Mean \emph{(left)} and median \emph{(right)} loss over different initializations.
  }
% \subfigure
\label{fig: learning dynamics}
\end{figure}
Finally we can now analyze in the next section how such a resolution can be used in the context of machine learning to improve convergence speed of learning trajectories.
Using the above concepts as a starting point we can conduct experiments that extend the analysis of the cone model in \cite{SingularDynamics} to gradient decent on the previously derived smooth manifold approximation. We focus our analysis on the convergence speed and form of the learning trajectories. 
To do so, we recall the two problems that can arise in the context of singularities: (1) slower convergence as result from attractor behaviour \emph{near the singularity}. 
(2)  the tangent space \emph{at the singularity} is undefined.

To analyze the behavior near and at the singularity, we consider \emph{gradient decent} over different initialization. Formally let $\bm\theta^{(t)}$ be the model parameters at time-step $t$, then the update rule for GD is defined as
$
\bm\theta^{(t+1)}=\bm\theta^{(t)}-\vartheta\nabla_{\bm\theta}\gL_c(x ; \bm\theta^{(t)})$
where $\vartheta$ is the learning rate and $\nabla_{\bm\theta}\gL_c(x ; \bm\theta^{(t)})$ the gradient of the negative log-likelihood $\gL(x ; \bm\theta)$ with respect to the parameters $\bm\theta^{(t)}$. 
Furthermore, the NGD update rule is given by
$\bm\theta^{(t+1)}=\bm\theta^{(t)}-\vartheta\gI^{-1}(\theta^{(t)}) \nabla_{\bm\theta}\gL(x ; \xi, \theta^{(t)},\epsilon)$
where  $\gI({\bm\theta})$ is the Fisher information matrix (FIM). For readability we omit the indication of the time-step in the following.
 
Applying these natural/ordinary GDs to the double cone model, we can write
$\gL_{c}(x ; \bm \mu):=\frac{1}{2}(\sum_{i=1}^3 (x_{i}-\mu_i)^{2})$ with the following reparameterization
$\bm \mu:=(\mu_1, \mu_2, \mu_3) := (\xi,
     \sqrt{\xi^2+\epsilon}\cos{\theta},
     \sqrt{\xi^2+\epsilon}\sin{\theta}).$
From there it is easy to obtain the learning dynamics under GD with $\overline{x}_i = \mathbb{E}[x_i]$ using
\begin{align*}%\label{eq: cone SGD}
    \nabla_x\gL_c=
     \begin{bmatrix}
     \overline{x}_1 + \overline{x}_2\cos\theta^{(t)}+\overline{x}_3\sin\theta^{(t)}- 2\xi^{(t)}\\
     -\xi^{(t)} (\overline{x}_2\sin\theta^{(t)}+\overline{x}_3\cos\theta^{(t)})
     \end{bmatrix},
     \quad
     \gI_{\mathrm{cone}} = \begin{bmatrix} 2&0\\0&\xi^2\end{bmatrix}.
\end{align*}
Similarly for NGD we obtain the learning dynamics for the smooth manifold approximation of the cone model by first defining the reparameterisation $(\widetilde\mu_1, \widetilde\mu_2, \widetilde\mu_3) = (\xi,
     \sqrt{\xi^2+\epsilon}\cos{\theta},
     \sqrt{\xi^2+\epsilon}\sin{\theta})$
so that we obtain the following gradient and FIM:
\begin{align*}%\label{eq: stratifold SGD}
    \nabla_x\gL_{\mathrm{hyp}}=
     \begin{bmatrix}
     -\overline{x}_1 +2\xi - \xi\overline{x}_2\cos \theta\frac{1}{\sqrt{\xi^2 + \epsilon}}
    - \xi\overline{x}_3\sin \theta\frac{1}{\sqrt{\xi^2 + \epsilon}}
     \\
    \sqrt{\xi^2 + \epsilon}\overline{x}_2\sin\theta
    -\sqrt{\xi^2 + \epsilon}\overline{x}_3\cos\theta
     \end{bmatrix}
     ,\quad
  \gI_{hyp}=\begin{bmatrix} \frac{\epsilon + 2\xi^2}{\epsilon+\xi^2}&0\\0&\epsilon + \xi^2\end{bmatrix}.
\end{align*}

Now using the above update equations, we obtain the learning trajectories  and loss curves as illustrated in Figure~\ref{fig: learning dynamics}. As we are interested in the comparison of GD on the initial model and the hyperboloid we note that for parameters that are initialized close to the true parameterization, the behavior for both settings is very similar as it does not come close to the singularity. In contrast if we focus on the learning dynamic plotted in blue, GD provides an update that gets trapped at the singularity and results in a plateau behaviour. On the other hand if we look at the same initialization on the hyperboloid, the influence of the singularity is less pronounced; the model does not get trapped and converges faster.

Interestingly if we consider \emph{NGD}, the convergence speed on both the cone model and the hyperboloid is very similar. This first might seem to be surprising however previous results \cite{SingularDynamics} showed that if we account for the FIM, the singularity has less influence on the learning trajectory.

\begin{figure}[t!]
	\centering
	\includegraphics[width=0.75\linewidth]{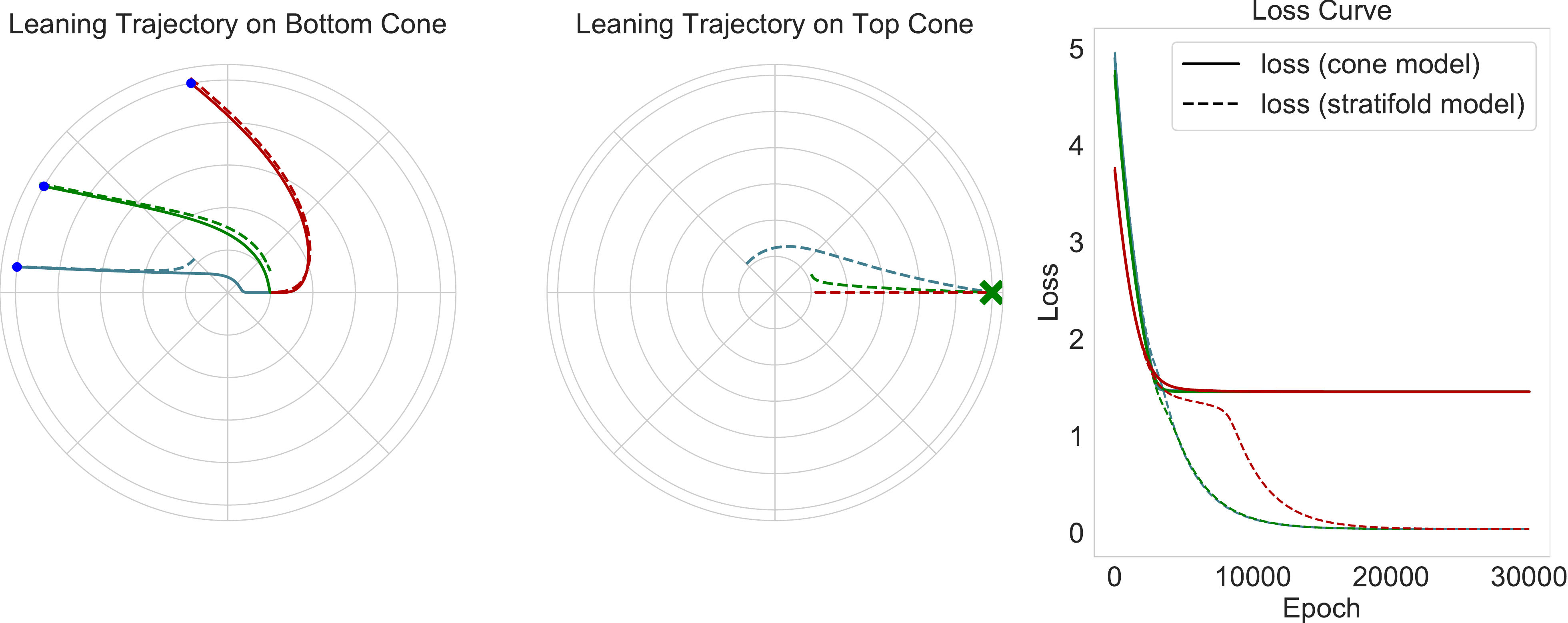}
	\caption[Initialization on  $\xi >0$ and target on  $\xi <0$ (different cones)]
	{Comparing loss and learning trajectory where the initialization and true parameterization are on different cones. The parameters have to go through the singularity as $\xi = 0$.
	\emph{(left)} cone one $\xi >0$.
	\emph{(middle)} cone two $\xi <0$.
	\emph{(right)} loss function.
}
	\label{fig: two cones}
\end{figure}%

This immediately brings up the question ``\textit{Why are we interested in the resolution of the model if the NGD seems to give a fast convergence?}'' Answering this brings us to the the second question posed in the introduction: \emph{what is the behavior of the learning dynamics at the singularity?}
We note that the Fisher information may potentially be infinite when the integral diverges: For example, this happens for the parametric family of uniform distributions $\{U[a,a+1]\ : \ a\in\mathbb{R}\}$.
In that case, Amari~\cite{Amari-1984} proposed to use Finsler geometry instead of Riemannian geometry by defining the information using Hellinger divergence between infinitesimally close distributions. 
Furthermore we observe that the FIM degenerates at the singularity, therefore problems arise when we consider a setting where we have to go \emph{through} the singularity (as illustrated in Figure~\ref{fig: two cones}). 
In the case of the cone model this happens when we initialize the parameters on one cone while the true parameters lie on the second cone. We indeed observe that the parameters get stuck at the singularity and we can not reach the true parameterization on the second cone. In contrast we see that for the hyperboloid, while we see a slowdown around $\xi = 0$, GD reaches the true parameterization. Therefore even in a NGD setting the consideration of the resolution can play a valuable role. 
Finally on an application level, we note that the use of NGD also comes with limitations \cite{martens2020new} like the requirement to computation of the inverse of the FIM which can become computationally intensive for complex models.

%%%
\section{Conclusion and Future Work}
%%%

We take a first step towards constructing a general framework for modeling  geometrically singular parameter spaces. By taking advantage of the nature of Stratifolds, we are able to describe parameter spaces that do not fulfill smoothness assumptions and use the resolution of this description of the parameter space  to obtain a smooth manifold approximation of the singular space.  Experimentally, we show that using first-order methods (SGD) on the smooth manifold, approximation decreases the effect of the singularity and therefore speeds up learning convergence. For natural gradient descent, we additionally find that the Fisher information metric can no longer degenerate during training as the resolution offers an approximation at previously singular points.

As we usually consider extensions to vanilla first-order gradient-based optimization like \cite{Adam,RMS,Adagrad} or optimizations that take the parameter surface into account \cite{RiemannianGradient,RiemannianGradient2} as a step into considering the parameter space more explicitly, those approaches do not require explicit knowledge of the topological properties of the parameter space. While first steps are taken in this direction \cite{Lin2021ICML}, a main focus of future research will be to develop a better understanding of the topology of the parameter spaces of commonly used models and formally characterize their singular points and their resolution. While the presented toy examples would allow for simpler geometric constructions, we strongly expect that more complex structures such as Stratifolds will be necessary to describe more complex models.

In line with the this direction, we can recall that for this paper, we focused on the very simple case of isolated singularities. It is to be expected that studying the topology of parameter spaces of complex models will also reveal more complex singularities. While \cite{ResolutionOS} provides first theoretical results on the resolution of more complex singularities, it is open how applicable those results are in practice.
Nevertheless, we conclude by emphasizing that the goal of this paper is  to gain a better understanding of how the structure of the parameter space influences the learning behavior, and propose the idea of resolving singular parameter spaces into smooth manifolds.

\vskip 0.3cm
\noindent {Acknowledgments.} This work was carried out while Pascal Esser was doing an internship at Sony Computer Science Laboratories in Tokyo, Japan.

% \clearpage
\printbibliography

\clearpage
\appendix

\section{Notations and Illustration of Blowup}

In this section we first give some notes on the general notations as well as some additional illustrations for concepts used in the main paper.

\subsection{Notations} \label{sec: notation and setup}

We denote algebras by $\Ca$, differential spaces as $(X,\Ca)$, and tangent spaces as $T_xX$ with the differential map $d f_{x} : T_{x} X \rightarrow T_{f(x)} X^{\prime}$. Manifolds are given by $\gM,\gN$ and Stratifolds by $\St$. For a topological space $\gS$, we denote the interior by $\stackrel{\circ}{\mathbf{\gS}}$ and the boundary as $\partial\gS$.
The germ of a function at a point describes how the function behaves very close to the point. For a point $x \in X$, we consider the germs of functions at $x, \bm{C}_x$.
For readability, we overload the notion of \emph{resolution} as the function mapping from a singular space onto the smooth manifold as well as the manifold resulting from the resolution itself.

\subsection{Additional Illustrations}

\begin{figure}[b]
    \centering
    \includegraphics[width=0.85\linewidth]{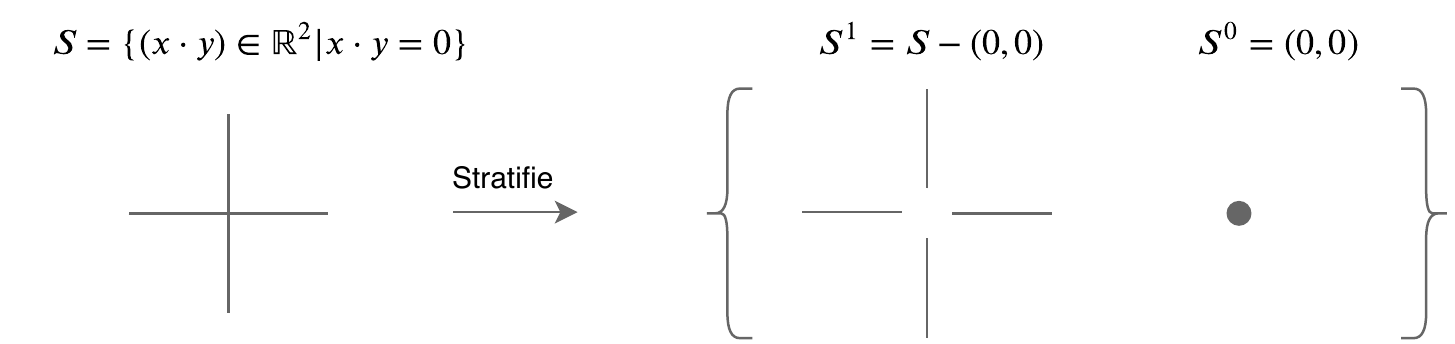}
    \caption[Algebraic Varietie $x\cdot y =0$ as Stratifold]{A simple example for a Stratifold. Considering a simple algebraic variety $x\cdot y =0$ with a stratification as described above.}
    \label{fig:xyStratifold}
\end{figure}

\begin{itemize}
    \item In Section~\ref{sec: Introduction}, we discussed the comparison between the resolution as used in \cite{watanabe_2009} and this paper. We further illustrate this in Figure~\ref{fig:blowup}.
    
    \item 	A standard n-simplex  is the subset of $\sR^{n+1}$ given by:
	\begin{align*}
	\Delta^{n}=\left\{\left(t_{0}, \ldots, t_{n}\right) \in \mathbb{R}^{n+1}\  \middle|\ \sum_{i=0}^{n} t_{i}=1, t_{i} \geq 0\forall i\right\}
	\end{align*}
	For an illustration of the inductive construction see Figure~\ref{fig: simplex}.

    \item The most natural examples of manifolds with singularities occur in algebraic geometry as algebraic varieties, i.e., zero sets of a family of polynomials. This example is mentioned in the main part and illustrated in Figure~\ref{fig:xyStratifold}.
\end{itemize}

\clearpage
\section{Additional Definitions: An Introduction to Topology} \label{app: sec: additional definitions 1}

% \section{An Introduction to Topology} \label{sec: topology}

In this appendix, we will recap some of the fundamental concepts from Topology, specifically related to smooth manifolds, as those concepts are essential for the later definition of Stratifolds.
The specific notation and phrasing of the definitions of this appendix are taken from \cite{ResolutionOS} but are not meant as a comprehensive introduction but as a reference for clarifying the terminology in the main part.

\subsection{Differential Space}
\begin{mydef}[Algebra]\label{def: algebra}
Denote the set of \textbf{continuous functions} from $X\rightarrow\mathbb{R}$ by $C^0(X)$ 
		
A subset $\bm C \subset C^0(X)$ is called an \textbf{algebra} of continuous functions if for $f,g \in \bm C$ the sum $f+g$ the product $fg$ and all constant functions  are in $\bm C$.
\end{mydef}{}

\begin{myex}
	set of functions $f : U \rightarrow \mathbb{R}$, where \textbf{all partial derivatives} of all orders exist, given by $C^\infty(U)$
\end{myex}{}

\begin{mydef}[Locally Detectable]
	Let $\mC$ be a subalgebra of the algebra of continuous functions $f : X \rightarrow\sR$. 
	
	We say that $\mC$ is \textbf{locally detectable} if  a function $h : X\rightarrow \sR$ is contained in $\mC$ if and only if for all $x \in X$ there is an open neighbourhood $U$ of $x$ and $g\in\mC$ such that $h|_U =g|_U$.
\end{mydef}{}

\begin{mydef}[Differential Space]
A \textbf{differential space} is a pair $(\bm X, \mC)$, where $\bm X$ is a topological space and $\Ca \subset C^0(X)$ is a locally detectable subalgebra of the algebra of continuous functions  satisfying the condition:
	
	For all $f_1,...,f_k \in \Ca$ and smooth functions $g : \sR^k \rightarrow \sR$, the function 
	\begin{align*}
	x \mapsto g\left(f_{1}(x), \ldots, f_{k}(x)\right)
	\end{align*}{}
	is in $\Ca$.
\end{mydef}{}

\subsection{Smooth Manifolds}

\begin{mydef}[homeomorphism]
A function $f:X\to Y$ between two topological spaces is a homeomorphism if it has the following properties:
\begin{itemize}
\item $f$  is a bijection
\item $f$ is continuous and 
\item the inverse function $f^{-1}$  is continuous 
\end{itemize}

\end{mydef}

% \begin{myex}
% A continuous deformation between a coffee mug and a donut (torus) illustrating that they are homeomorphic. See \autoref{fig:mug and donut}.
% \begin{figure}[h!]
% 	\centering
% 	\includegraphics[width=0.5\linewidth]{drawio/Stuff-Donut.pdf}
% 	\caption[Homeomorphism]{}
% 	\label{fig:mug and donut}
% \end{figure}%
% \end{myex}
\begin{myex}
A chart of a manifold is an homeomorphism between an open subset of the manifold and an open subset of a Euclidean space.
\end{myex}

\begin{myex}
$\sR^m$ and $\sR^n$ are not homeomorphic for $m \neq n$. (no bijaction)
\end{myex}

\begin{mydef}[Isomorphism]
	Let $(X,\Ca)$ and $(X',\Ca')$ be differential spaces. A homeomorphism $f : X \rightarrow X'$ is called an isomorphism if for each $g \in \Ca'$ and $h\in\Ca$, we have $gf\in\Ca$ and $hf^{-1} \in\Ca'$.
\end{mydef}{}

\begin{mydef}[smooth manifold]
	A $k$-dimensional smooth manifold is a \textbf{differential space} $(M, \Ca)$ where $M$ is a Hausdorff space with a countable basis of its topology, such that for each $x \in M$ there is an open neighbourhood $U \subseteq M$, an open subset $V \subset \sR^k$ and an \textbf{isomorphism}
	\begin{align*}
	\varphi :\left(V, C^{\infty}(V)\right) \rightarrow(U, \mathbf{C}(U)).
	\end{align*}
	a $k$-dimensional smooth manifold is a differential space which is locally isomorphic to $\sR^k$
\end{mydef}{}

\begin{mydef}[Germ]
The germ of a function at a point describes how the \textbf{function behaves very close to the point}%, where "very close" allows us to consider an arbitrarily small open subset containing the point.
		
For a point $x \in X$, we consider the germs of functions at $x, \Ca_x$. If $f \in \Ca$ and $g \in \Ca$ are representatives of germs at $x$, then the sum $f + g$ and the product $f \cdot g$ represent well-defined germs denoted $[f]_x + [g]_x \in \Ca_x$ and $[f]_x \cdot [g]_x \in \Ca_x$.
\end{mydef}{}

\begin{mydef}[Derivation]
	Let $(X, \Ca)$ be a differential space. A derivation at $x \in X$ is a map from the germs of functions at $x$
	\begin{align*}
	\alpha : \mathbf{C}_{x} \longrightarrow \mathbb{R}
	\end{align*}
	s.t.
	\begin{align*}
	\alpha\left([f]_{x}+[g]_{x}\right)&=\alpha\left([f]_{x}\right)+\alpha\left([g]_{x}\right)\\
	\alpha\left([f]_{x} \cdot[g]_{x}\right)&=\alpha\left([f]_{x}\right) \cdot g(x)+f(x) \cdot \alpha\left([g]_{x}\right)\\
	\alpha\left([c]_{x} \cdot[f]_{x}\right)&=c \cdot \alpha\left([f]_{x}\right)    
	\end{align*}{}
	for all $f,g \in \Ca$ and $[c]_x$ the germ of the constant function which maps all $y \in X$ to $c \in \sR$.
\end{mydef}{}

\begin{mydef}[Tangent Space]
	Let $(X,\Ca)$ be a differential space and $x \in X$. The \textbf{vector space of derivations} at $x$ is called the tangent space of $X$ at $x$ and denoted by $T_xX$.
\end{mydef}{}

\begin{mydef}[Differential]
	Let $f : (X, \Ca) \rightarrow (X', \Ca')$ be a morphism. Then for each $x \in X$ the differential
	\begin{align*}
	d f_{x} : T_{x} X \rightarrow T_{f(x)} X^{\prime}
	\end{align*}
	is the map which sends a derivation $\alpha$ to $\alpha'$ where $\alpha'$ assigns to $[g]_{f(x)} \in \Ca'_{x'}$
	the value $\alpha([gf]_x)$.
\end{mydef}{}

\begin{figure}[h!]
	\centering
	\includegraphics[width=0.8\linewidth]{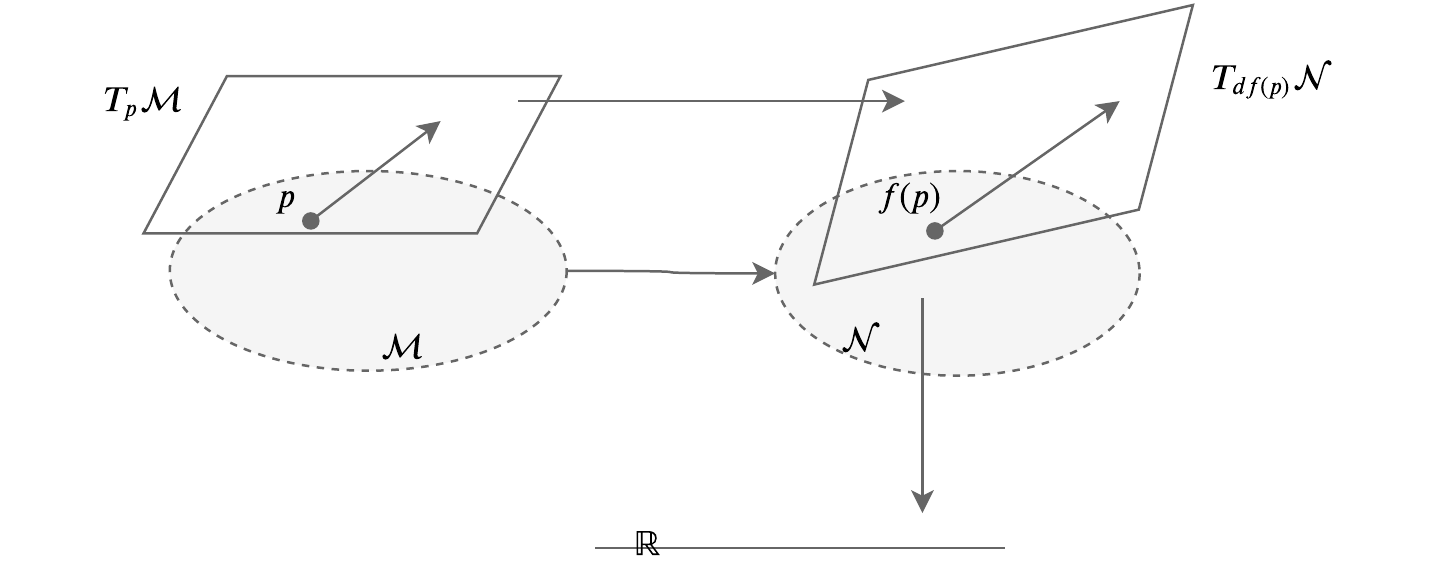}
	\caption[Illustration of a Differential Map]{Illustration of a Differential map $	d f_{x} : T_{x} X \rightarrow T_{f(x)} X^{\prime}$. Consider $\gM$ and $\gN$ be smooth manifolds, displayed are only a part of it \footnotemark}
	\label{fig:differential map}
\end{figure}%
\footnotetext{As the illustration might be misleading: An important note here is that at $p$ we have to to be able to define the tangent space $ T_{p} \gM$. Therefore the point can not lie on the boundary of a topological space. We will further discuss this in the following, Section~\ref{sec: boundaries}.}

\subsection{Boundaries}\label{sec: boundaries}

Besides the very standard definitions presented above, we have to consider some concepts that are related 

Let $(\St,\partial\St)$  be a pair of topological spaces. We denote $\St-\partial\St$ by $\Sti$ and call it the \textbf{interior}. We assume that $\Sti$ and $\partial\St$ are stratifolds of dimension $n$ and $n - 1$ and that $\partial\St$ is a closed subspace.

\begin{figure}[h!]
	\centering
	\includegraphics[width=0.5\linewidth]{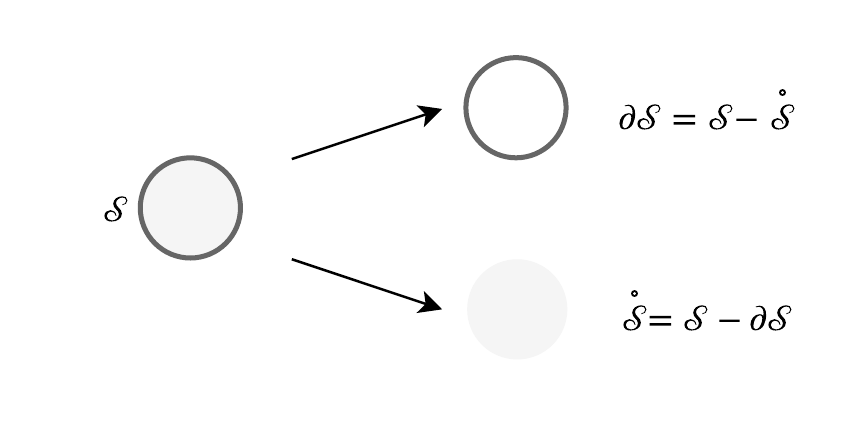}
	\caption[Illustration of a Topological Boundary]{boundary: For a topological space we consider the \textbf{interior} and \textbf{boundary} or the space, as illustrated above}
	\label{fig:Collor}
\end{figure}%

Now we want to look at the map onto a neighborhood of the boundary:

\begin{mydef}[Collar]
	Let $(\St,\partial\St)$ be a pair as above. A collar is a homeomorphism
	\begin{align*}
	\mathbf{c} : U_{\epsilon} \rightarrow V
	\end{align*}
	where $\epsilon>0, U_{\epsilon} :=\partial \St \times[0, \epsilon)$ and $V$ is an open neighbourhood of $\partial\St$ in $\St$ such that $\mathbf{c} |_{ \partial \St \times\{0\}}$ is the identity map to $\partial\St$ and $\mathbf{c} |_{U_\varepsilon- (\partial \St \times\{0\})}$ is an isomorphism of stratifolds onto $V-\partial\St$
	
	Intuitively: the boundary cross an interval open on one side, equivalently a small open neighborhood of the boundary.
\end{mydef}{}

\clearpage
\section{More Formal setup of Section~\ref{sec: Introduction to stratifolds}}\label{app: sec: additional definitions 2}

While in Section~\ref{sec: Introduction to stratifolds} we generally refer to Stratifolds, we note that there are technical more details to be considered. Therefore we state the following definitions.

\begin{mydef}[$k$-dimensional Stratifold \cite{StratifoldsBook}]
    
    A $k$-dimensional Stratifold is a differential space $(\St, \Ca)$, where S is a locally compact    Hausdorff space    with countable basis, and the skeleta $\Sigma^i$ are closed subspaces. In addition we assume:
    \begin{enumerate}
        \item Restriction gives a smooth structure on $\St^i$ and for each $x \in \St^i$ restriction gives an isomorphism
        \[i^{*} : \mathbf{C}_{x} \stackrel{\cong}{\rightarrow} C^{\infty}\left(\St^{i}\right)_{x}\]
        
        \item  All tangent spaces have dimension $\leq k$,
    \[\operatorname{dim} T_{x} \St \leq k \ \ \forall \ x \in \St\]
        
        \item (Bump Function) for each $x \in \St$ and open neighbourhood $U \subset \St$ there is a non-negative function $\rho \in \Ca$ such that $\rho(x) \neq 0 $ and $\text{supp } \rho \subseteq U$
    \end{enumerate}{}
\end{mydef}{}

A natural examples of spaces with singularities occur in algebraic geometry as algebraic varieties, i.e., zero sets of a family of polynomials.
To get a better intuition, consider the following example: an algebraic variety of the form $x\cdot y =0$ as Stratifold. Consider the parameter space defined over $\{x,y\}\text{ s.t. }x\cdot y =0$ we can stratify this space into a $0$-dimensional manifold as $\{0,0\}$ and four axis elements.
In addition we define the following specific class of Stratifolds.

\begin{mydef}[$c$-stratifold \cite{StratifoldsBook}]
    An $n$-dimensional $c$-stratifold $\St$ (a collared Stratifold) is a
    pair of topological spaces $(\St,\partial\St)$  together with a germ of collars $[\bm c]$ where 
    $\St = \Sti-\partial\St$ is an $n$-dimensional Stratifold and $\partial\St$ is an $(n-1)$-dimensional Stratifold, which is a closed subspace of $\St$. We call $\partial\St$  the boundary of $\St$.
    
    A smooth map from $\St$ to a smooth manifold $\gM$ is a continuous function $f$ who's restriction to $\stackrel{\circ}{\mathbf{\St}}$ and to $\partial\St$ is smooth and which commutes with an appropriate representative of the germ of collars, i.e., there is a $\gamma>0$ such that $f\bm c(x,t) = f(x)$ for all $x \in \partial\St$ and $t < \delta$.
    
\end{mydef}{}

Those refined definitions are not central for the simple example considered in the main paper but will play a central role when transferring the idea to more complex models.

\begin{myrem}[Uniqueness of the decomposition]
We can easily construct examples for the same stratification but different Stratifolds as well as one Space different Stratifolds.
\end{myrem}

For the main paper we consider the simple case of an isolated singularly as defined below. Again for more complex models an extension to not singular singularities will be necessary. \cite{ResolutionOS} provides a theoretical framework for such cases but we formalize the isolated case below.

\begin{mydef}[Isolated Singularity]
    Let $\{x\}_{i\in\sN}$ be a countable set of points and let  $g:\partial\gM\rightarrow\{x\}_{i\in\sN}$ be a a proper map from an $m$-dimensional smooth manifold $\gM$ onto a set of a $0$-dimensional manifold. Writing it as a Stratifold  gives us 
\begin{align*}
    \St = \gM \cup_g \lc x_i\rc_i.
\end{align*}{}
A $m$-dimensional Stratifold $\St$ is said to have isolated singularity iff 
\begin{align*}
    \St^i = \emptyset \quad \forall i\in \lc 1, \cdots, m-1\rc
\end{align*}{}
\end{mydef}

\subsection{Resolution of Algebraic Varieties with Isolates Singularities \cite{ResolutionOS}}

Consider an algebraic variety $V\subset \mathbb{R}^n$ with isolated singularities $s_i\in \Sigma$ where $\Sigma$ is the singular set, all points  $\lc s_i\rc_i$ are 0-dimensional manifold.
In the case where $s_i$ is open in $V$ we do not need to resolve the point.

Therefore consider the other option. Let $s_i$ be closed in $V$ for the following. From there we define a distance function $\rho_i(x): = ||x-s_i||^2$ on $\sR^n$ and let $D_{\varepsilon_i}(s_i)$ be the enclosing ball in $\mathbb{R}^n$ with radius $\varepsilon_i$ around $s_i$.
Now the idea is that we define an enclosing ball around the singularity and define a union of the initial algebraic variety and the enclosing ball. We can define this as
    \begin{align*}
    \partial V_{\varepsilon_i}(s_i):=V_{\varepsilon_i}(s_i)\cap \partial D_{\varepsilon_i}(s_i)
    \end{align*}
such that the restriction $\rho_i|_{ V_{\varepsilon_i} - \{s_i\}}$ has no critical value. With that we can define a continuous map
    \begin{align*}
    \overline{h}: \partial V_{\varepsilon_i}(s_i)\times [0,\varepsilon_i]\rightarrow V_{\varepsilon_i}
    \end{align*}
this gives us a c-manifold $W$ with obvious collar over the enclosing ball:
    \begin{align*}
    W :=V-\left( \bigsqcup
    _{i} \mathring{D}_{\varepsilon_{i}}\left(s_{i}\right)\right) \cup_{\mathrm{id}} \partial V_{\varepsilon_{i}}\left(s_{i}\right) \times\left[0, \varepsilon_{i}\right]
    \end{align*}
 finally the map
    \begin{align*}
    f = \text{id} \cup \overline{h}: W\rightarrow \mathcal{V}
    \end{align*}
    gives $\mathcal{V}$ the structure of a Stratifold with isolated singularities

\subsection{Resolution of Hypersurfaces \cite{ResolutionOS}}
Again we  assume an algebraic variety, a polynomial $p:\sR^{n+1}\rightarrow\sR$ with singularity $\{s_i\}_i$
\[
s_i \in V:=p^{-1}(0).
\]
We are interested in the link to the singularity $\partial V_{\varepsilon_i}(s_i)$.
Choose a $\delta>0$ such that for all $c$ with $|c| \leq \delta$ are regular values of $p$ and pick $c$ such that
\[
p^{-1}(c)\neq\emptyset
\]
then $p^{-1}(c)$ is a smooth manifold. In the case of a complex polynomial $p:\mathbb{C}^{n+1}\rightarrow\mathbb{C}(n>0)$, every deformation $p^{-1}(c)$ gives an optima resolution given $||c||$ is small enough \cite{ResolutionOS}.

\begin{myrem}
    In the above definitions and also the resolution considered afterward, we only consider a specific class of Stratifolds. To be specific, we have to distinguish between \textit{p-stratifolds and c-stratifolds} as defined in \cite{ResolutionOS}.
    For a rigorous definition of the Stratifolds spaces and the resolution, those distinctions are technically important as they define exact restrictions on allowed mappings and the properties of the space. 
     In particular, Stratifolds constructed inductively by attaching manifolds together using the data: germs of collars and attaching maps, are called parameterized Stratifolds or p-stratifolds \cite{StratifoldsBook}.
    For this paper and to illustrate the main ideas we omit those distinctions there but note that revisiting them will become important when modeling more complex models.

\end{myrem}{}

\end{document}